\documentclass[pdflatex,sn-mathphys-num]{sn-jnl}

\usepackage{graphicx}%
\usepackage{multirow}%
\usepackage{amsmath,amssymb,amsfonts}%
\usepackage{amsthm}%
\usepackage{mathrsfs}%
\usepackage[title]{appendix}%
\usepackage{xcolor}%
\usepackage{textcomp}%
\usepackage{manyfoot}%
\usepackage{booktabs}%
\usepackage{algorithm}%
\usepackage{algorithmicx}%
\usepackage{algpseudocode}%
\usepackage{listings}%
\usepackage{makecell}
\usepackage{booktabs}
\usepackage{amsmath, amssymb}
\usepackage{supertabular}
\usepackage{longtable}
\usepackage{tcolorbox}
\usepackage{makecell}
\usepackage{caption}
\tcbuselibrary{listings, skins, breakable} 

\lstdefinestyle{PythonStyle}{
    language=Python,
    captionpos=b,
    basicstyle=\ttfamily\small,
    keywordstyle=\color{blue},
    stringstyle=\color{orange},
    commentstyle=\color{green!60!black},
    identifierstyle=\color{black},
    showstringspaces=false,
    numbers=left,
    numberstyle=\tiny\color{gray},
    frame=single,
    breaklines=true
}

\theoremstyle{thmstyleone}%
\theoremstyle{thmstyletwo}%

\theoremstyle{thmstylethree}%

\begin{document}

\title[Article Title]{The Sound of Populism: Distinct Linguistic Features Across Populist Variants}


\author[1]{\fnm{Yu} \sur{Wang}}
\author[2]{\fnm{Runxi} \sur{Yu}}
\author[1]{\fnm{Zhongyuan} \sur{Wang}}
\author*[3]{\fnm{Jing} \sur{He}}\email{jinghe@fudan.edu.cn}

\affil[1]{\orgdiv{Fudan Institute for Advanced Study in Social Sciences}, \orgname{Fudan University}, \orgaddress{\city{Shanghai}, \postcode{200433}, \country{China}}}

\affil[2]{\orgname{YK Pao School}, \orgaddress{\city{Shanghai}, \postcode{201620}, \country{China}}}

\affil[3]{\orgdiv{College of Foreign Languages and Literatures}, \orgname{Fudan University}, \orgaddress{\city{Shanghai}, \postcode{200433}, \country{China}}}



\abstract{This study explores the sound of populism by integrating classic Linguistic Inquiry and Word Count (LIWC) features, which capture the emotional and stylistic tones of language, with a fine-tuned RoBERTa model, a state-of-the-art context-aware language model trained to detect nuanced expressions of populism. This approach allows us to uncover the auditory dimensions of political rhetoric in U.S. presidential inaugural and State of the Union addresses. We examine how four key populist dimensions (i.e., left-wing, right-wing, anti-elitism, and people-centrism) manifest in the linguistic markers of speech, drawing attention to both commonalities and distinct tonal shifts across these variants. Our findings reveal that populist rhetoric consistently features a direct, assertive ``sound" that forges a connection with ``the people'' and constructs a charismatic leadership persona. However, this sound is not simply informal but strategically calibrated. Notably, right-wing populism and people-centrism exhibit a more emotionally charged discourse, resonating with themes of identity, grievance, and crisis, in contrast to the relatively restrained emotional tones of left-wing and anti-elitist expressions.
}

\keywords{Populism, Fine-Tuning, Large Language Models, LIWC, Text as Data, Computational Social Science}




\maketitle



\section{Introduction}

Populism is defined both as an ideology and a discursive strategy.
It is a thin ideology that considers society to be ultimately separated
into two homogeneous and antagonistic groups, ``the pure people'' versus
``the corrupt elite,'' and which argues that politics should be an expression
of the general will of the people~\citep{muddecasPopulistZeitgeist2004}.
It is also taken ``as a discursive strategy that juxtaposes the virtuous
populace with a corrupt elite and views the former as the sole legitimate
source of political power''~\citep{bonikowskiPopulistStyleAmerican2016}. Laclau characterizes populism not as a fixed ideology but
as a discourse that constructs political identities and relations~\citep{laclauPopulistReason2005}. In the process of economic
and socio-cultural grievances being articulated across social strata,
a collective vague and broad identity of the ``people'', as opposed
to ``elites'', is rhetorically constructed.
Populism is observable across various political landscapes globally, from left-wing social movements and right-wing nationalist campaigns.
Its surge in recent years was evident in events like Brexit, the rise
of AfD, the outsider political candidates in the U.S. and Argentina,
especially the twice election of Donald Trump. Besides its wide political
and geographical coverage, populism also spans chronologically. Presidents
like Andrew Jackson and Ronald Reagan occasionally employed ``us
vs. them'' rhetoric that can be seen as early populist echoes. Populism
on one hand is understood as a correction for elite-dominated politics
and a moral attitude since it ``identifies the will of the people
with justice and morality''~\citep{shilsTormentSecrecyBackground1956}.
On the other hand, it is perceived as a malaise of democracy, where
public opinion is manipulated and hijacked by a few opinion leaders
through resentment, fear, and anger to serve their own political agenda~\citep{PopulistAppeals}.

In presidential rhetoric, language is a primary tool for shaping political
reality and public opinion. U.S. presidents, through inaugural addresses
and State of the Union (SOTU) speeches, not only communicate policies
but also construct narratives about national identity, values, and
who holds power. Prior research shows that changes in linguistic style
can reflect broader socio-political trends \citep{jordanExaminingLongtermTrends2019,tuckerDataScienceApproach2020}.
For example, the gradual but consistent decline in the readability
index of congressional speeches after the 1970s can be possibly attributed
to the increase in the broadcasting of speeches through the media
where ``legislators started to address the media and the general
public through their speeches.''\citep{tuckerDataScienceApproach2020}
Studying the linguistic features of presidential speeches, therefore,
is not just a linguistic exercise but a window into how leaders frame
``the people'' and ``the elite,'' legitimize their authority,
and persuade audiences.

Our study leverages two widely used techniques --- Linguistic Inquiry
and Word Count (LIWC) \citep{Liwc}
and fine-tuned RoBERTa-large model \citep{roberta}---
to analyze the ``sound'' of populism in U.S. presidential speeches: how different flavors of populism are reflected in potentially different language styles. LIWC is a text analysis tool that counts words in psychologically relevant
categories (e.g., pronouns, emotion expressions, cognitive processes), providing
a quantitative linguistic profile of a document. It has been widely
used to profile political language. For instance, LIWC-style analyses
have revealed trends like declining analytical complexity in presidents'
speeches over time and shifts in emotional tone \citep{tuckerDataScienceApproach2020}.
RoBERTa, on the other hand, is a context-aware language model that
can be fine-tuned to detect subtle patterns indicative of populism
beyond simple word counts. Recent advances show that transformer models
can reliably classify texts by ideological leaning or populist content,
matching human evaluations with high accuracy \citep{Erhard_Hanke_Remer_Falenska_Heiberger_2025}. 

Coupling LIWC's interpretable linguistic features with RoBERTa's pattern-recognition prowess, we analyze the full corpus of U.S. presidential
inaugural and SOTU speeches to pinpoint the linguistic
markers that characterize the different flavors of populism. We take the manifestation of
\textbf{populism} in these addresses as the dependent variable and the \textbf{linguistic features} extracted via LIWC (to
capture the psychological and stylistic profile of the speech) as
the key independent variables. By modeling populism as a function of LIWC features, we investigate the following key question:


\begin{itemize}
\item What specific linguistic patterns are characteristic of each of the four distinct subtypes of populism?
\end{itemize}

To preview our findings, we find common linguistic features of different dimensions of populism discourses. Three dimensions, left-wing, right-wing and anti-elitism, embrace informality and reject complexity, non-fluency and tentativeness. This direct and assertive rhetoric style helps to build a closer bond with the people and project a confident image for the populist leader. However, both left-wing and anti-elitism show negative coefficients with swearing, which means that they maintain discursive legitimacy by avoiding vulgarity. The informality and authenticity projected by populist discourse are calibrated. We also find cleavage among different dimensions of populism. Right-wing and people-centrism are more emotionally charged to build a moral narrative of crisis and identity, with people-centrism highlights strong associations with social and economic references to build a narrative of economic injustice, while left-wing and anti-elitism focus more on structural reform with less emotional appeal. 
Overall, populist rhetoric across dimensions employs linguistic styles strategically to craft clear, confident, and emotionally resonant messages that emphasize certainty, collective identity, and opposition to elites, while maintaining ideological distinctions.

\section{Literature Review}

\label{literature-review}
Numerous empirical studies have explored the nature of populism in political communication, revealing both its content and its distinctive tonal characteristics. \citet{bonikowskiPopulistStyleAmerican2016} demonstrated that populist rhetoric appears not only among fringe candidates but also across both major U.S. political parties. Outsider or insurgent candidates are particularly reliant on populist claims, often positioning themselves as champions of the “ordinary citizen” against the political establishment. This trend is echoed by \citet{Hawkins2018}, who observed a surge in populist discourse during periods of political disruption and noted an increase in populist rhetoric globally in the 21st century, particularly in Europe and Latin America.

\subsection{Four Flavors of Populism}

\label{theoretical-foundations-of-populism}

\citet{Erhard_Hanke_Remer_Falenska_Heiberger_2025} detect populism on four dimensions: anti-elitism, people-centrism, left-wing ideology, and right-wing ideology. \citet{muddecasPopulistZeitgeist2004}'s definition of populism, as quoted in the Introduction, encapsulates a Manichaean ``us'' versus ``them'' worldview, and a dichotomy between good and evil. The adversarial discourse style that highlights the antagonistic relationship between the ``us, the people'' and the ``them, the elites'' and a moral struggle between the ``virtuous'' and the ``corrupt'', which are distinct from typical technocratic speeches or partisan debate, both formulates and manifests the ideological kernel of populism: anti-elitism and people-centrism.

Anti-elitism perceives elites as corrupt, self-serving, and disconnected from the concerns of the common people. Elites can be politicians, business leaders, media figures, or any dominant group seen as exploiting or ignoring the interests of the majority. Donald Trump’s 2016 presidential campaign used the slogan ``Drain the Swamp'' \citep{newyorker2016drainingswamp} to symbolize his anti-elitist stance. Trump positioned himself as an outsider seeking to cleanse Washington D.C. of corrupt politicians and special interests, appealing to the disillusioned masses by promising to clamp down on lobbying, reduce corporate influence, and implement term limits for lawmakers~\citep{Hawkins2018}.

People-centrism emphasizes the ``people'' as a unified, virtuous population, in contrast with corrupt elites, as the true source of legitimacy and moral authority in politics. It often entails appeals to the ``common sense'' of ordinary citizens and portrays politics as a struggle between the pure people and the corrupt elite. For example, the ``Leave'' Campaign of Brexit argued that the ``people'' should take back control from EU elites \citep{takebackcontrol}.

The concept of people is fluid rather than fixed. It can be divided on an economic basis like working class or underprivileged citizens, or formed on a cultural line as a nation or ethnicity \citep{Kriesi2014,Canovan2002}. When it is grounded in economic justice and class struggle, populism manifests itself as left-wing populism. It typically advocates social equality, economic redistribution, labor rights, unionization and opposition to capitalism and neoliberal policies \citep{mudde2017populism}. While it shares populism's anti-elitism, it specifically targets corporate elites, multinational institutions, and neoliberal politicians rather than cultural or ethnic outsiders. For example, Hugo Chávez's rhetoric opposed global capitalism and U.S. imperialism while advocating for wealth redistribution and social welfare policies \citep{Simon2023}. While not a full-fledged populist, Bernie Sanders uses populist rhetoric against billionaires and corporate greed, emphasizing the need for Medicare for All and Fair wages \citep{sunkara2019socialist}.

Right-wing populism is rooted in nationalism, cultural conservatism, opposition to immigration, and an emphasis on traditional values \citep{Mudde2004}. Right-wing populists tend to portray the "elite" as liberal globalists, intellectuals, or bureaucrats who undermine national identity and sovereignty. Anti-immigration rhetoric claims immigrants take jobs, undermine culture, or increase crime. Donald Trump’s ``America First" rhetoric attacks political elites, the media, and immigrants while appealing to nationalism \citep{Hawkins2018}. Marine Le Pen’s National Rally promotes nationalism, anti-immigration policies, and opposition to the European Union \citep{chrisafis2024}.

While both left- and right-wing populists oppose elites, left-wing populists blame economic elites (corporations, banks), while right-wing populists often blame intellectuals, globalists, media, immigrants, and are more focused on cultural or racial issues. 

\subsection{Linguistic Features of Populism Rhetoric: Insights from Prior Studies and LIWC Analysis}

A growing body of research has examined the linguistic features of populist rhetoric, shedding light on how populist politicians employ distinctive strategies in their communication.\citet{nai2021fear} found that populist politicians, across 40 countries, tended to employ a more negative tone, using fear-based language 8 percent more than mainstream politicians. \citet{tuckerDataScienceApproach2020} demonstrated that Donald Trump’s speeches during the 2016 U.S. election campaign were delivered at a lower reading level than those of his political peers. This linguistic simplicity is often interpreted as a strategy to connect with the “common people,” who are portrayed as less educated \citep{mudde2017populism,laclauPopulistReason2005,Hawkins2018}. However, this view has been challenged by \citet{zanottoLanguageComplexityPopulist2020}, who found that readability scores did not consistently distinguish populist from non-populist speeches across various contexts. A more recent example is that Trump’s 2025 inaugural address was notably policy-intensive, deviating not only from his previous speeches but also from the typical rhetorical style of populist discourse \citep{trump2025inaugural}. This suggests that populists may adjust their linguistic strategies depending on the situational context. These findings imply that populism is not solely about content—such as the division between ‘the people’ and ‘the elite’—but also about tone. Populist rhetoric is frequently characterized by a sharply polarizing emotional style, often centered on grievance and distrust. Additionally, it tends to be simpler and more colloquial compared to the language of elite figures.

 LIWC is a tool designed to investigate the psychological and social dynamics embedded in discourses, as well as language complexity and accessibility. It not only captures articles, prepositions, and auxiliary verbs to analyze syntactic complexity, but also provides a quantifiable profile of a text’s emotional and cognitive underpinnings by categorizing words into dimensions such as “Emotional Tone,” “Cognition,” and “Social” factors. Categories such as “Analytical Thinking,” “Clout,” and “Authenticity” are particularly relevant for understanding populism, as they help to measure the rhetorical strategies used by populist leaders to construct an image of directness, authenticity, and authority~\citep{pennebakerDevelopmentPsychometricProperties2015}. LIWC’s ability to identify linguistic complexity and patterns in the emotional tone, cognitive framing, and social dynamics of political speech makes it an invaluable tool for empirically assessing the “sound” of populism and its psychological resonance with audiences

\label{linguistic-features-as-indicators-of-populism}

\subsection{Measuring Populist Flavors with BERT Models}

\label{identifying-ideological-tendencies-with-bert-and-modern-nlp}

While dictionaries and manual coding have advanced our understanding on populism~\citep{rooduijn2011measuring,populist_ideas_on_social_media}, recent developments in natural language processing (NLP) offer newer and more powerful means to detect and measure ideological tendencies like populism~\citep{doi:10.1177/00491241221122317}. Bidirectional
Encoder Representations from Transformers (BERT) is a language model that has achieved state-of-the-art
results in text classification tasks due to its ability to understand
context and nuanced language patterns. In the realm of political and sociological text analysis, researchers have begun fine-tuning BERT models on labeled data to classify texts by ideology, partisanship, topic, and populist content~\citep{pretrained_topic_classification,finetune_pa,discrete_emotions}.~\citet{doi:10.1177/00491241221122317} fine-tuned a RoBERTa model~\citep{roberta} to measure populism, nationalism, and authoritarianism in U.S. campaign speeches spanning from 1952 (Dwight Eisenhower vs. Adlai Stevenson II) to 2020 (Donald Trump vs. Joe Biden). \citet{Erhard_Hanke_Remer_Falenska_Heiberger_2025} fine-tuned a German BERT model~\citep{gbert}, which is pretrained on German text, to detect populism in speeches delivered in the German parliament. Extending beyond a general conceptualization of populism, they examined its subtypes by measuring texts along four dimensions: anti-elitism, people-centrism, left-wing populism, and right-wing populism.



Our research builds on the burgeoning literature on populism detection by integrating LIWC and BERT models, combining BERT's superior performance in measuring populism with the interpretability of LIWC-derived features~\citep{political_research_exchange}.

\section{Data and Methods}
\subsection{Data}
Our study draws on two primary data sources: U.S. presidential speeches and German parliamentary speeches.
 
We collected the full text of U.S. presidential speeches from the American Presidency Project\footnote{\url{https://www.presidency.ucsb.edu/documents/presidential-documents-archive-guidebook/annual-messages-congress-the-state-the-union}}, a comprehensive archive of presidential documents including inaugural addresses, executive orders, State of the Union (SOTU) addresses, and more. For this study, we focus specifically on inaugural addresses (1789–2025) and State of the Union addresses (1790–2025), given their rhetorical and political significance. The final corpus consists of 308 speeches. Table~\ref{liwc} in the Appendix presents the summary statistics for the LIWC-derived features extracted from these texts. In the following, we highlight illustrative excerpts from speeches by Andrew Jackson (1829) and Donald Trump (2017).
\begin{tcolorbox}[colback=blue!15,colframe=black!50!white,title=Speech Snippet by Andrew Jackson in 1829]
“As long as our Government is administered for the good of the people, and is regulated by their will; as long as it secures to us the rights of person and of property, liberty of conscience and of the press, it will be worth defending; and so long as it is worth defending a patriotic militia will cover it with an impenetrable aegis. Partial injuries and occasional mortifications we may be subjected to, but a million of armed freemen, possessed of the means of war, can never be conquered by a foreign foe. To any just system, therefore, calculated to strengthen this natural safeguard of the country I shall cheerfully lend all the aid in my power."
\end{tcolorbox}

\begin{tcolorbox}[colback=red!15,colframe=black!50!white,title=Speech Snippet by Donald Trump in 2017]
“Washington flourished – but the people did not share in its wealth.\\

Politicians prospered – but the jobs left, and the factories closed.\\

The establishment protected itself, but not the citizens of our country.\\

Their victories have not been your victories; their triumphs have not been your triumphs; and while they celebrated in our nation’s Capital, there was little to celebrate for struggling families all across our land."
\end{tcolorbox}












For the German parliamentarian speeches, we utilize the dataset from~\citet{Erhard_Hanke_Remer_Falenska_Heiberger_2025}. The dataset contains 8,795 sentences from German parliamentarian debates from 2013 to 2021. Each sentence is annotated by five annotators for four different labels: left-wing, right-wing, anti-elitism, and people-centrism. During the aggregation stage, a sentence is coded as positive (1) for a given label if at least one annotator assigns that label to the sentence. Table~\ref{snippets} presents illustrative speech snippets, each annotated with four labels.

\begin{table}[ht]
\begin{tabular}{|p{5.5cm}|c|c|c|c|}
\hline
\textbf{German Text (English translation)} & \textbf{Anti-elitism} & \textbf{People-centrism} & \textbf{Left} & \textbf{Right} \\ \hline
Wir kriegen das nicht hin, wenn nur wir in Deutschland dafür unsere Industrie schreddern. (We won't be able to do this if we in Germany are the only ones who shred our industry.
) & \makecell{0} & 0 & 0 & 0 \\ \hline
Sie bitten wieder die Normalbürger zur Kasse und schonen die fetten Konten. (They are once again asking ordinary citizens to pay and protecting their fat bank accounts.
) & 1 & 1 & 1 & 0 \\ \hline
Wir bedauern das Aussterben der Deutschen durch die kinder- und familienfeindliche Politik der jetzigen Bundesregierung. (We regret the extinction of the German population due to the anti-child and anti-family policies of the current federal government.) & 
1 & 0 & 0 & 1 \\ \hline
Da gibt es den Lkw-Fahrer, der den Gammelfleischskandal aufdeckt, oder die couragierte Pflegefachkraft, die die Missstände in ihrer Einrichtung mutig anprangert. (There is the truck driver who uncovers the rotten meat scandal, or the courageous nurse who bravely denounces the abuses in her facility.) & 0 & 1 & 0 & 0 \\ \hline
\end{tabular}
\caption{Sample texts from the German dataset~\citep{Erhard_Hanke_Remer_Falenska_Heiberger_2025}. Each text is labeled for anti-elitism, people-centrism, left-wing, and right-wing.}
\label{snippets}
\end{table}

\subsection{Methods}
\subsubsection{LIWC for Independent Variables}
To operationalize linguistic style and psychological content, we extracted a comprehensive set of features from each speech segment using the Linguistic Inquiry and Word Count (LIWC) lexicon. Table~\ref{liwc} in the Appendix presents the summary statistics (mean, standard deviation, minimum, quartiles, and maximum) for each of 94 LIWC variables across the corpus of 308 speech segments, which includes U.S. presidential inaugural addresses and State of the Union addresses. These variables encompass syntactic features (e.g., function words, pronouns), cognitive and emotional processes (e.g., insight, anger, anxiety), and stylistic dimensions (e.g., tone, authenticity, informality). The descriptive statistics provide a foundational overview of the linguistic landscape used as independent variables in our subsequent regression analyses.

\subsubsection{Transformer Models for Dependent Variables}

As baselines, we use German BERT~\citep{gbert} for multi-label classification.\footnote{\citet{Erhard_Hanke_Remer_Falenska_Heiberger_2025} fine-tune a German BERT on this dataset using cross-validation. However, the cross-validation step is not reproducible.} German BERT has 336 million parameters. It accepts German as input and is fine-tuned to generate four scores between 0 and 1, representing left-wing, right-wing, anti-elitism, and people-centrism, respectively. For batch size and weight decay, we adopt the same parameter values as reported in~\citet{Erhard_Hanke_Remer_Falenska_Heiberger_2025} and~\citet{finetune_pa}. The key hyperparameter we tune is the learning rate, where we explore values of 15e-6, 10e-6, and 5e-6. German BERT is fine-tuned for 3 epochs~\citep{Erhard_Hanke_Remer_Falenska_Heiberger_2025}. Instead of reporting results from a single run, we average the outcomes over three random runs. For each run, we select the optimal checkpoint based on the macro-averaged F1 score on the validation set.\\

We further fine-tune a RoBERTa-large model~\citep{roberta}, which is comparable in size to BERT~\citep{bert} and GBERT~\citep{gbert}, using English translations of the German dataset. RoBERTa was chosen for its strong performance among models in the BERT family~\citep{chatgpt_vs_bert,model_selection}. The fine-tuning setup mirrors that of German BERT, including the number of training epochs and batch size, with the only difference being our learning rate search space (2e-5, 1.75e-5, and 1.5e-5).
Instead of using the original German text, we translated the snippets into English using GPT-4o.\footnote{Translations were performed via the OpenAI API. Interested readers could refer to Table~\ref{google} in the Appendix for our results using Google Translate.}

\begin{table}[h!]
\centering
\setlength{\tabcolsep}{10pt} 
\begin{tabular}{|l|c|c|c|c|c|c|}
\hline
 & \multicolumn{3}{c|}{\citet{Erhard_Hanke_Remer_Falenska_Heiberger_2025}(baseline)} & \multicolumn{3}{c|}{Ours} \\\hline
 \textbf{Dimension} & \textbf{Precision} & \textbf{Recall} & \textbf{F1} & \textbf{Precision} & \textbf{Recall} & \textbf{F1} \\ \hline
Anti-elitism                        & 0.81               & 0.88            & 0.84        & 0.82               & 0.89            & \textbf{0.85}        \\ \hline
People-centrism                     & 0.70               & 0.73            & 0.71        & 0.66               & 0.78            & \textbf{0.71}        \\ \hline
Left-wing ideology                  & 0.69               & 0.77            & \textbf{0.73}        & 0.73               & 0.72            & 0.72        \\ \hline
Right-wing ideology                 & 0.68               & 0.66            & 0.67        & 0.68               & 0.73            & \textbf{0.70}        \\ \hline
\textbf{micro average}                  & 0.75               & 0.80            & 0.77        & 0.75               & 0.81            & \textbf{0.78}        \\ \hline
\textbf{macro average}                  & 0.72               & 0.76            & 0.74        & 0.72               & 0.78            & \textbf{0.75 }       \\ \hline
\end{tabular}
\caption{Performance metrics across different dimensions from~\cite{Erhard_Hanke_Remer_Falenska_Heiberger_2025} and our new benchmark. Our benchmark is the average of three random runs. Better results (higher F1 score) in bold.}
\label{tab:metrics}
\end{table}

\section{Experiments}
To examine how linguistic features vary across different subtypes of populism, we regress the populism scores (e.g., left-wing, people-centrism) on the linguistic features extracted using LIWC. Specifically, we employ a multiple linear regression framework, where \( Y \) denotes the score of a populism subtype and \( X_1, X_2, \ldots, X_p \) represent the LIWC features, and Year denotes the year of the speech (Equation 1).

\begin{equation}
Y = \beta_0 + \beta_1 X_1 + \beta_2 X_2 + \cdots + \beta_p X_p + \gamma Year + \varepsilon
\end{equation}

In the following, we present our regression results and discussion. Given the large number of LIWC features, we focus exclusively on the significant coefficients with a \textit{p}-value less than 0.05.

\subsection*{Regression Results: Left-wing populism}
\

\begin{table}[!h]
\setlength{\tabcolsep}{20pt} 
\centering
\begin{tabular}{lrr}
\toprule
 & Coefficient & p-value \\
\midrule
WC & 0.0000 & 0.0450 \\
anx & -0.0926 & 0.0400 \\
anger & -0.0833 & 0.0171 \\
focuspresent & -0.0286 & 0.0393 \\
informal & 2.8163 & 0.0000 \\
swear & -2.7297 & 0.0000 \\
netspeak & -2.3629 & 0.0000 \\
assent & -2.8047 & 0.0000 \\
nonflu & -2.9681 & 0.0000 \\
\bottomrule
\end{tabular}
\caption{Among the LIWC features with statistically significant coefficients for left-wing populism, both informal language and word count are positively associated with left-wing populist discourse.}
\end{table}

\noindent We find strong positive association with informal language aligns with left-wing populism's ``bottom-up'' discursive style~\citep{laclauPopulistReason2005}, which attempts to sound closer to ``the people'' and distant from elites or technocrats. This finding aligns with \citet{laclauPopulistReason2005}'s populist logic , where constructing a collective populist identity relies on accessible, inclusive rhetoric.

The negative coefficients on swear confirm \citet{Wodak2015}'s observation that left-wing populism tends to use inclusive  and civil language to build coalition narratives. 

 The negative coefficients on netspeak, and nonfluency testify \citet{Moffitt2016}'s  ``calibrated authenticity'' theory that while left-wing populist actors often adopt stylized performances of informality and "bottom-up" style to seem authentic, its informality is often calibrated and strategic, not chaotic since a high nonfluency score may signal indecision, lack of clarity, or lack of conviction — all of which undermine populism’s assertive, binary, and moralistic tone. This reflects a more ideologically driven and more structured messages.

Anger and anxiety are also negatively associated with left-wing populism, suggesting left populist discourse may emphasize hopeful transformation or systemic critique more than emotional outbursts. This fits \citet{judis2016populist}'s differentiation of left from right populism: the left-wing tends to focus on economic injustice and structural reform, whereas the right is more emotionally reactive (e.g., anger at immigrants or elites). 

\subsection*{Regression Results: Right-wing populism}

\begin{table}[!h]
\setlength{\tabcolsep}{20pt} 
\centering
\begin{tabular}{lrr}
\toprule
 & Coefficient & p-value \\
\midrule
article & -0.0148 & 0.0196 \\
prep & -0.0153 & 0.0051 \\
auxverb & -0.0136 & 0.0403 \\
adverb & -0.0144 & 0.0108 \\
tentat & -0.0144 & 0.0160 \\
percept & -0.0352 & 0.0462 \\
feel & 0.0583 & 0.0052 \\
ingest & -0.0733 & 0.0320 \\
power & 0.0134 & 0.0305 \\
Parenth & 0.2545 & 0.0351 \\
\bottomrule
\end{tabular}
\caption{Among the LIWC features with statistically significant coefficients for right-wing populism, both \textit{feel} and \textit{power} are positively associated with right-wing populist discourse.}
\end{table}

\noindent For right-wing populism, the negative coefficients on articles, prepositions, auxiliary verbs, adverbs manifest that its language is less syntactically complex. These lexicons are typically used to structure more grammatically sophisticated sentences \citep{lenormandRoleFunctionWords2022}. \citet{Lakoff1990} and \citet{pennebaker2003psychological} suggest that syntactic complexity is a proxy for cognitive complexity. Populist leaders on the right wing appear to use simpler structures to project clarity and strength.

The reduced use of tentative language and perception-related terms, combined with the positive associations with ‘feel’ and ‘power,’ aligns with \citet{Hawkins2018}’s theory that right-wing populism employs an authoritarian discourse style and charismatic leadership narratives. These narratives are grounded in dominance and rely more on emotional appeals than logical arguments, favoring a binary worldview where actors are seen as either entirely good or evil. Such rhetoric is often emotionally and morally charged, with little regard for nuance, particularly in discussions of nationalism, sovereignty, and fear of societal decline \citep{Corina}.

\subsection*{Regression Results: Anti-elitism}

Anti-elitism strongly correlates with informality, consistent with its moral dichotomy between  “pure people vs. corrupt elite” narrative \citep{Mudde2004}. Informality serves to challenge hierarchical discourse, delegitimize elite authority and create ideological distance from elite norms.

\begin{table}[htbp]
\setlength{\tabcolsep}{20pt} 
\centering
\begin{tabular}{lrr}
\toprule
 & Coefficient & p-value \\
\midrule
death & -0.1852 & 0.0339 \\
informal & 3.3700 & 0.0036 \\
swear & -5.0262 & 0.0002 \\
netspeak & -3.1472 & 0.0102 \\
assent & -3.2690 & 0.0051 \\
nonflu & -3.2723 & 0.0060 \\
\bottomrule
\end{tabular}
\caption{Among the LIWC features with statistically significant coefficients for anti-elitism, \textit{informal} language shows a positive association, while \textit{swear} and \textit{netspeak} among others are negatively associated.}
\end{table}

The strong negative coefficients for swear and netspeak suggest that, similar to left populism, anti-elitist messages are controlled and purposeful, avoiding descending into vulgarity by erratic or subcultural slang to maintain discursive legitimacy while still appearing informal and emotionally resonant. This suggests a strategic populism rather than merely venting anger.

The negative coefficient on assent shows that anti-elitism minimizes hedges or polite markers like ``I agree" ``yes". This finding implies that anti-elitist discourse is assertive, not cooperative, which is consistent with \citet{EuropeanPopulism2015} that anti-elitist narratives thrive on discursive certainty, promoting an unambiguous opposition to the elite rather than engaging in cooperative or conciliatory speech. As \citet{hawkins2009ischavez} notes, “The populist leader claims to know the will of the people—not to guess it” .

Similar with left-wing, anti-elitism is negatively associated with nonflu, showing that while populist leaders often reject elite jargon, they also project rhetorical control to channel a collective identity. Disfluency might suggest that the speaker is struggling to articulate that voice, which could disrupt the illusion of unity and clarity, which is especially important for anti-elitism when it depends on combative tone and moral clarity to project the dichotomy between ``us'' and ``them.'' Therefore, both left-wing and anti-elitist populists must walk a tightrope: informal but assertive, relatable but confident.

\subsection*{Regression Results: People-centrism}
In Table~\ref{people_centrism} we report the results on people centrism. We observe that use of questions is negatively associated with people-centrism. Populist rhetoric is often assertive rather than inquisitive, projecting a singular voice of the people~\citep{hawkins2009ischavez}. We also note that the coefficient on posemo is negative and statistically significant, suggesting that people-centrist language often carry less positive emotions. This is consistent with the argument that populist and people-centric messages often carry grievance, urgency, or criticism, rather than optimism~\citep{10.1093/sf/sov120}. Positive emotion language may undercut the ``us vs. them'' dichotomy central to populism~\citep{mudde2017populism}.

\begin{table}[!htbp]
\centering
\setlength{\tabcolsep}{20pt} 
\begin{tabular}{lrr}
\toprule
\toprule
Variable & Coefficient & p-value \\
\midrule
interrog & -0.2485 & 0.0255 \\
posemo & -1.1276 & 0.0460 \\
social & 0.1818 & 0.0424 \\
cause & -0.2940 & 0.0489 \\
tentat & -0.3737 & 0.0038 \\
money & 0.1923 & 0.0024 \\
\bottomrule
\end{tabular}
\caption{Among the LIWC features with statistically significant coefficients for people-centrism, \textit{social} and \textit{money} are positively associated, while \textit{interrog} and \textit{posemo} exhibit negative associations.}\label{people_centrism}
\end{table}

One variable, social, refers to words such as talk, share and friend. Its coefficient is positively associated with people-centrism. Populist discourse frequently uses social referents to construct the ``people'' as a collective identity~\citep{hawkins2009ischavez}. Social words help create a sense of in-group cohesion, consistent with people-centrism. Furthermore, we note that references to money are also positively associated with people-centrism. According to~\citet{norris2019cultural}, economic grievances are central to many populist narratives, especially those that claim the elites have stolen wealth from the people. In our analysis, this is particularly so for people-centrism rhetoric.

\section{Limitations and Future Work}
While this research captures the nuanced tones across different variants of populism, several limitations must be acknowledged. 

First, the analysis relies heavily on U.S. presidential inaugural and SOTU addresses and German parliamentary speeches, which may not fully represent the full spectrum of populist communications where populism may manifest in distinct linguistic forms. Future research could expand the corpus to include populist speeches from other democracies~\citep{gpd}, as well as other settings, such as campaign speeches~\citep{politics_as_usual}, where
political leaders engage with their audiences in less formal and more spontaneous contexts and their language could be even more emotionally charged and/or strategically calibrated. Expanding the dataset to include discourses of such broader spectrum would provide a more comprehensive and comparative view of how populist rhetoric operates across different settings.

Second, although the combination of LIWC and transformer models offers a more granular approach than traditional manual coding, it still faces challenges in fully capturing the subtleties of human emotion conveyed in political rhetoric. LIWC’s psycholinguistic categories, while widely used, are not exhaustive, and more intricate emotional dynamics and some nuances of populist discourse, such as rhetorical tropes, irony, and sarcasm, may not be adequately captured. 

\section{Conclusion} 
This study offers a nuanced examination of the linguistic features inherent in different variants of populist discourse, specifically across left-wing, right-wing, anti-elitism, and people-centrism. By integrating the Linguistic Inquiry and Word Count (LIWC) tool with a fine-tuned RoBERTa language model, we uncover distinct patterns in U.S. presidential rhetoric, emphasizing both the tonal and ideological variances within populist language.

Our findings affirm that populist discourse, irrespective of its ideological alignment, adopts a direct, assertive communication style, which seeks to establish a strong connection with “the people” while distinguishing populist leaders from elite institutions. However, the strategies for constructing this connection differ significantly across populist dimensions. Right-wing populism and people-centrism are characterized by emotionally charged narratives, invoking themes of crisis, identity, and grievance. In contrast, left-wing populism and anti-elitism emphasize structural reform and critique, using less emotionally intense rhetoric to maintain focus on economic and social justice.

Furthermore, the study demonstrates that populist rhetoric strategically balances informality with control, rejecting complexity and nonfluency in favor of clarity and decisiveness. This balance helps maintain populism’s moral and ideological clarity, even as it rejects elite discourse. Notably, the avoidance of vulgarity in left-wing and anti-elitist populism further underscores the strategic calibration of tone to preserve rhetorical legitimacy while still fostering a sense of authenticity.

These insights provide a deeper understanding of how populist leaders utilize language not merely as a tool of persuasion, but as a means of shaping collective identity and consolidating political power. By refining our analytical methods and extending the application of LIWC and transformer-based models, this research contributes to a more sophisticated framework for studying populism, offering valuable insights into its linguistic underpinnings and the emotional resonance it evokes in contemporary political discourse.

\section*{Data and Code Availability}
The data and code used in this study will be made publicly available upon acceptance of the manuscript.

\section*{Competing Interests}
The author(s) declare that they have no competing interests.









\newpage
\bibliography{x}


\begin{thebibliography}{47}
\ifx \bisbn   \undefined \def \bisbn  #1{ISBN #1}\fi
\ifx \binits  \undefined \def \binits#1{#1}\fi
\ifx \bauthor  \undefined \def \bauthor#1{#1}\fi
\ifx \batitle  \undefined \def \batitle#1{#1}\fi
\ifx \bjtitle  \undefined \def \bjtitle#1{#1}\fi
\ifx \bvolume  \undefined \def \bvolume#1{\textbf{#1}}\fi
\ifx \byear  \undefined \def \byear#1{#1}\fi
\ifx \bissue  \undefined \def \bissue#1{#1}\fi
\ifx \bfpage  \undefined \def \bfpage#1{#1}\fi
\ifx \blpage  \undefined \def \blpage #1{#1}\fi
\ifx \burl  \undefined \def \burl#1{\textsf{#1}}\fi
\ifx \doiurl  \undefined \def \doiurl#1{\url{https://doi.org/#1}}\fi
\ifx \betal  \undefined \def \betal{\textit{et al.}}\fi
\ifx \binstitute  \undefined \def \binstitute#1{#1}\fi
\ifx \binstitutionaled  \undefined \def \binstitutionaled#1{#1}\fi
\ifx \bctitle  \undefined \def \bctitle#1{#1}\fi
\ifx \beditor  \undefined \def \beditor#1{#1}\fi
\ifx \bpublisher  \undefined \def \bpublisher#1{#1}\fi
\ifx \bbtitle  \undefined \def \bbtitle#1{#1}\fi
\ifx \bedition  \undefined \def \bedition#1{#1}\fi
\ifx \bseriesno  \undefined \def \bseriesno#1{#1}\fi
\ifx \blocation  \undefined \def \blocation#1{#1}\fi
\ifx \bsertitle  \undefined \def \bsertitle#1{#1}\fi
\ifx \bsnm \undefined \def \bsnm#1{#1}\fi
\ifx \bsuffix \undefined \def \bsuffix#1{#1}\fi
\ifx \bparticle \undefined \def \bparticle#1{#1}\fi
\ifx \barticle \undefined \def \barticle#1{#1}\fi
\bibcommenthead
\ifx \bconfdate \undefined \def \bconfdate #1{#1}\fi
\ifx \botherref \undefined \def \botherref #1{#1}\fi
\ifx \url \undefined \def \url#1{\textsf{#1}}\fi
\ifx \bchapter \undefined \def \bchapter#1{#1}\fi
\ifx \bbook \undefined \def \bbook#1{#1}\fi
\ifx \bcomment \undefined \def \bcomment#1{#1}\fi
\ifx \oauthor \undefined \def \oauthor#1{#1}\fi
\ifx \citeauthoryear \undefined \def \citeauthoryear#1{#1}\fi
\ifx \endbibitem  \undefined \def \endbibitem {}\fi
\ifx \bconflocation  \undefined \def \bconflocation#1{#1}\fi
\ifx \arxivurl  \undefined \def \arxivurl#1{\textsf{#1}}\fi
\csname PreBibitemsHook\endcsname

\bibitem[\protect\citeauthoryear{Mudde}{2004}]{muddecasPopulistZeitgeist2004}
\begin{botherref}
\oauthor{\bsnm{Mudde}, \binits{C.}}:
The {{Populist Zeitgeist}}
\textbf{39}(4),
542--563
(2004)
\end{botherref}
\endbibitem

\bibitem[\protect\citeauthoryear{Bonikowski and Gidron}{}]{bonikowskiPopulistStyleAmerican2016}
\begin{botherref}
\oauthor{\bsnm{Bonikowski}, \binits{B.}},
\oauthor{\bsnm{Gidron}, \binits{N.}}:
The {{Populist Style}} in {{American Politics}}: {{Presidential Campaign Discourse}}, 1952–1996.
Social Forces
\textbf{94}(4),
1593--1621
\doiurl{10.1093/sf/sov120} .
Accessed 2025-03-02
\end{botherref}
\endbibitem

\bibitem[\protect\citeauthoryear{Laclau}{2005}]{laclauPopulistReason2005}
\begin{bbook}
\bauthor{\bsnm{Laclau}, \binits{E.}}:
\bbtitle{On Populist Reason}.
\bpublisher{Verso Books},
\blocation{London}
(\byear{2005})
\end{bbook}
\endbibitem

\bibitem[\protect\citeauthoryear{Shils}{1956}]{shilsTormentSecrecyBackground1956}
\begin{bbook}
\bauthor{\bsnm{Shils}, \binits{E.}}:
\bbtitle{The {{Torment}} of {{Secrecy}}. {{The Background}} and the {{Consequences}} of {{American Security Policies}}.}
\bpublisher{Free Press},
\blocation{Glencoe}
(\byear{1956})
\end{bbook}
\endbibitem

\bibitem[\protect\citeauthoryear{Aytaç et~al.}{2025}]{PopulistAppeals}
\begin{barticle}
\bauthor{\bsnm{Aytaç}, \binits{S.E.}},
\bauthor{\bsnm{Çarkoğlu}, \binits{A.}},
\bauthor{\bsnm{Elçi}, \binits{E.}}:
\batitle{Populist appeals, emotions, and political mobilization}.
\bjtitle{American Behavioral Scientist}
\bvolume{69}(\bissue{5}),
\bfpage{507}--\blpage{525}
(\byear{2025})
\doiurl{10.1177/00027642241240343}
{\href{https://arxiv.org/abs/https://doi.org/10.1177/00027642241240343}{{https://doi.org/10.1177/00027642241240343}}}
\end{barticle}
\endbibitem

\bibitem[\protect\citeauthoryear{Jordan et~al.}{2019}]{jordanExaminingLongtermTrends2019}
\begin{botherref}
\oauthor{\bsnm{Jordan}, \binits{K.N.}},
\oauthor{\bsnm{Sterling}, \binits{J.}},
\oauthor{\bsnm{Pennebaker}, \binits{J.W.}},
\oauthor{\bsnm{Boyd}, \binits{R.L.}}:
Examining long-term trends in politics and culture through language of political leaders and cultural institutions
\textbf{116}(9),
3476--3481
(2019)
\doiurl{10.1073/pnas.1811987116} .
Accessed 2025-03-01
\end{botherref}
\endbibitem

\bibitem[\protect\citeauthoryear{Tucker et~al.}{2020}]{tuckerDataScienceApproach2020}
\begin{botherref}
\oauthor{\bsnm{Tucker}, \binits{E.C.}},
\oauthor{\bsnm{Capps}, \binits{C.J.}},
\oauthor{\bsnm{Shamir}, \binits{L.}}:
A data science approach to 138 years of congressional speeches
\textbf{6}(8)
(2020)
\doiurl{10.1016/j.heliyon.2020.e04417}
{\href{https://arxiv.org/abs/32904137}{{32904137}}}.
Accessed 2025-03-01
\end{botherref}
\endbibitem

\bibitem[\protect\citeauthoryear{Pennebaker et~al.}{2015}]{Liwc}
\begin{botherref}
\oauthor{\bsnm{Pennebaker}, \binits{J.}},
\oauthor{\bsnm{Boyd}, \binits{R.}},
\oauthor{\bsnm{Jordan}, \binits{K.}},
\oauthor{\bsnm{Blackburn}, \binits{K.}}:
The Development and Psychometric Properties of LIWC2015,
Austin, TX: University of Texas at Austin
(2015).
\doiurl{10.15781/T29G6Z}
\end{botherref}
\endbibitem

\bibitem[\protect\citeauthoryear{Liu et~al.}{2019}]{roberta}
\begin{botherref}
\oauthor{\bsnm{Liu}, \binits{Y.}},
\oauthor{\bsnm{Ott}, \binits{M.}},
\oauthor{\bsnm{Goyal}, \binits{N.}},
\oauthor{\bsnm{Du}, \binits{J.}},
\oauthor{\bsnm{Joshi}, \binits{M.}},
\oauthor{\bsnm{Chen}, \binits{D.}},
\oauthor{\bsnm{Levy}, \binits{O.}},
\oauthor{\bsnm{Lewis}, \binits{M.}},
\oauthor{\bsnm{Zettlemoyer}, \binits{L.}},
\oauthor{\bsnm{Stoyanov}, \binits{V.}}:
{RoBERTa: A Robustly Optimized BERT Pretraining Approach}.
arXiv:1907.11692
(2019)
\end{botherref}
\endbibitem

\bibitem[\protect\citeauthoryear{Erhard et~al.}{2025}]{Erhard_Hanke_Remer_Falenska_Heiberger_2025}
\begin{botherref}
\oauthor{\bsnm{Erhard}, \binits{L.}},
\oauthor{\bsnm{Hanke}, \binits{S.}},
\oauthor{\bsnm{Remer}, \binits{U.}},
\oauthor{\bsnm{Falenska}, \binits{A.}},
\oauthor{\bsnm{Heiberger}, \binits{R.H.}}:
{{PopBERT}}. {{Detecting}} populism and its host ideologies in the german bundestag
\textbf{33}(1),
1--17
(2025)
\doiurl{10.1017/pan.2024.12}
\end{botherref}
\endbibitem

\bibitem[\protect\citeauthoryear{Hawkins et~al.}{2018}]{Hawkins2018}
\begin{bbook}
\bauthor{\bsnm{Hawkins}, \binits{K.A.}},
\bauthor{\bsnm{Carlin}, \binits{R.E.}},
\bauthor{\bsnm{Littvay}, \binits{L.}},
\bauthor{\bsnm{Kaltwasser}, \binits{C.R.}}:
\bbtitle{The Ideational Approach to Populism: Concept, Theory, and Analysis}.
\bpublisher{Routledge},
\blocation{New York}
(\byear{2018}).
\doiurl{10.4324/9781315196923} .
\burl{https://doi.org/10.4324/9781315196923}
\end{bbook}
\endbibitem

\bibitem[\protect\citeauthoryear{Widmer}{2017}]{newyorker2016drainingswamp}
\begin{botherref}
\oauthor{\bsnm{Widmer}, \binits{T.}}:
Draining the swamp.
The New Yorker
(2017).
Accessed: 2025-05-08
\end{botherref}
\endbibitem

\bibitem[\protect\citeauthoryear{Haughton}{2017}]{takebackcontrol}
\begin{barticle}
\bauthor{\bsnm{Haughton}, \binits{T.}}:
\batitle{Editorial: Take back control: The european union in 2016}.
\bjtitle{Journal of Common Market Studies}
(\byear{2017})
\doiurl{10.1111/jcms.12617}
\end{barticle}
\endbibitem

\bibitem[\protect\citeauthoryear{Kriesi}{2014}]{Kriesi2014}
\begin{barticle}
\bauthor{\bsnm{Kriesi}, \binits{H.}}:
\batitle{The populist challenge}.
\bjtitle{West European Politics}
\bvolume{37}(\bissue{2}),
\bfpage{361}--\blpage{378}
(\byear{2014})
\doiurl{10.1080/01402382.2014.887879}
\end{barticle}
\endbibitem

\bibitem[\protect\citeauthoryear{Canovan}{2002}]{Canovan2002}
\begin{bbook}
\bauthor{\bsnm{Canovan}, \binits{M.}}:
In: \beditor{\bsnm{Mény}, \binits{Y.}},
\beditor{\bsnm{Surel}, \binits{Y.}} (eds.)
\bbtitle{Taking Politics to the People: Populism as the Ideology of Democracy},
pp. \bfpage{25}--\blpage{44}.
\bpublisher{Palgrave Macmillan},
\blocation{New York, NY}
(\byear{2002}).
\doiurl{10.1057/9781403920072_2}
\end{bbook}
\endbibitem

\bibitem[\protect\citeauthoryear{Mudde and Rovira~Kaltwasser}{2017}]{mudde2017populism}
\begin{bbook}
\bauthor{\bsnm{Mudde}, \binits{C.}},
\bauthor{\bsnm{Rovira~Kaltwasser}, \binits{C.}}:
\bbtitle{Populism: A Very Short Introduction}.
\bsertitle{Very Short Introductions}.
\bpublisher{Oxford University Press},
\blocation{New York}
(\byear{2017}).
\doiurl{10.1093/actrade/9780190234874.001.0001} .
\bcomment{Online edition, Oxford Academic, 23 Feb. 2017. Accessed 28 Mar. 2025}.
\burl{https://doi.org/10.1093/actrade/9780190234874.001.0001}
\end{bbook}
\endbibitem

\bibitem[\protect\citeauthoryear{Simon and Parody}{2023}]{Simon2023}
\begin{barticle}
\bauthor{\bsnm{Simon}, \binits{J.}},
\bauthor{\bsnm{Parody}, \binits{G.}}:
\batitle{The devil and democracy in the global south: Hugo chávez's transnational populism}.
\bjtitle{Journal of Latin American Studies}
\bvolume{55}(\bissue{4}),
\bfpage{653}--\blpage{677}
(\byear{2023})
\doiurl{10.1017/S0022216X23000731}
\end{barticle}
\endbibitem

\bibitem[\protect\citeauthoryear{Sunkara}{2019}]{sunkara2019socialist}
\begin{bbook}
\bauthor{\bsnm{Sunkara}, \binits{B.}}:
\bbtitle{The Socialist Manifesto: The Case for Radical Politics in an Era of Extreme Inequality}.
\bpublisher{Basic Books},
\blocation{New York}
(\byear{2019}).
\burl{https://www.basicbooks.com/titles/bhaskar-sunkara/the-socialist-manifesto/9781541674004/}
\end{bbook}
\endbibitem

\bibitem[\protect\citeauthoryear{Mudde}{2004}]{Mudde2004}
\begin{barticle}
\bauthor{\bsnm{Mudde}, \binits{C.}}:
\batitle{The populist zeitgeist}.
\bjtitle{Government and Opposition}
\bvolume{39}(\bissue{4}),
\bfpage{541}--\blpage{563}
(\byear{2004})
\doiurl{10.1111/j.1477-7053.2004.00135.x}
\end{barticle}
\endbibitem

\bibitem[\protect\citeauthoryear{Chrisafis}{2024}]{chrisafis2024}
\begin{botherref}
\oauthor{\bsnm{Chrisafis}, \binits{A.}}:
On the brink of power: how france’s national rally reinvented itself.
The Guardian
(2024)
\end{botherref}
\endbibitem

\bibitem[\protect\citeauthoryear{Nai}{2021}]{nai2021fear}
\begin{botherref}
\oauthor{\bsnm{Nai}, \binits{A.}}:
Fear and loathing in populist campaigns? {{Comparing}} the communication style of populists and non-populists in elections worldwide
\textbf{20}(2),
219--250
(2021)
\end{botherref}
\endbibitem

\bibitem[\protect\citeauthoryear{Zanotto et~al.}{2020}]{zanottoLanguageComplexityPopulist2020}
\begin{botherref}
\oauthor{\bsnm{Zanotto}, \binits{S.E.}},
\oauthor{\bsnm{Frassinelli}, \binits{D.}},
\oauthor{\bsnm{Butt}, \binits{M.}}:
Language {{Complexity}} in {{Populist Rhetoric}},
61--80
(2020)
\end{botherref}
\endbibitem

\bibitem[\protect\citeauthoryear{Trump}{2025}]{trump2025inaugural}
\begin{botherref}
\oauthor{\bsnm{Trump}, \binits{D.J.}}:
The Inaugural Address.
\url{https://www.whitehouse.gov/remarks/2025/01/the-inaugural-address/}.
Accessed: 2025-05-08
(2025)
\end{botherref}
\endbibitem

\bibitem[\protect\citeauthoryear{Rooduijn and Pauwels}{2011}]{rooduijn2011measuring}
\begin{barticle}
\bauthor{\bsnm{Rooduijn}, \binits{M.}},
\bauthor{\bsnm{Pauwels}, \binits{T.}}:
\batitle{Measuring populism: Comparing two methods of content analysis}.
\bjtitle{West European Politics}
\bvolume{34}(\bissue{6}),
\bfpage{1272}--\blpage{1283}
(\byear{2011})
\end{barticle}
\endbibitem

\bibitem[\protect\citeauthoryear{Gründl}{2022}]{populist_ideas_on_social_media}
\begin{botherref}
\oauthor{\bsnm{Gründl}, \binits{J.}}:
Populist ideas on social media: A dictionary-based measurement of populist communication.
New Media \& Society
(2022)
\end{botherref}
\endbibitem

\bibitem[\protect\citeauthoryear{Bonikowski et~al.}{2022}]{doi:10.1177/00491241221122317}
\begin{barticle}
\bauthor{\bsnm{Bonikowski}, \binits{B.}},
\bauthor{\bsnm{Luo}, \binits{Y.}},
\bauthor{\bsnm{Stuhler}, \binits{O.}}:
\batitle{Politics as usual? measuring populism, nationalism, and authoritarianism in u.s. presidential campaigns (1952–2020) with neural language models}.
\bjtitle{Sociological Methods \& Research}
\bvolume{51}(\bissue{4}),
\bfpage{1721}--\blpage{1787}
(\byear{2022})
\doiurl{10.1177/00491241221122317}
{\href{https://arxiv.org/abs/https://doi.org/10.1177/00491241221122317}{{https://doi.org/10.1177/00491241221122317}}}
\end{barticle}
\endbibitem

\bibitem[\protect\citeauthoryear{Wang}{2023a}]{pretrained_topic_classification}
\begin{botherref}
\oauthor{\bsnm{Wang}, \binits{Y.}}:
Topic classification for political texts with pretrained language models.
Political Analysis
(2023)
\end{botherref}
\endbibitem

\bibitem[\protect\citeauthoryear{Wang}{2023b}]{finetune_pa}
\begin{botherref}
\oauthor{\bsnm{Wang}, \binits{Y.}}:
On finetuning large language models.
Political Analysis
(2023)
\end{botherref}
\endbibitem

\bibitem[\protect\citeauthoryear{Widmann}{2025}]{discrete_emotions}
\begin{botherref}
\oauthor{\bsnm{Widmann}, \binits{T.}}:
Do politicians appeal to discrete emotions? the effect of wind turbine construction on elite discourse.
Journal of Politics
(2025)
\end{botherref}
\endbibitem

\bibitem[\protect\citeauthoryear{Chan et~al.}{2020}]{gbert}
\begin{botherref}
\oauthor{\bsnm{Chan}, \binits{B.}},
\oauthor{\bsnm{Schweter}, \binits{S.}},
\oauthor{},
\oauthor{\bsnm{Möller}, \binits{T.}}:
German's next language model.
Proceedings of the 28th International Conference on Computational Linguistics
(2020)
\end{botherref}
\endbibitem

\bibitem[\protect\citeauthoryear{Hunger}{2024}]{political_research_exchange}
\begin{botherref}
\oauthor{\bsnm{Hunger}, \binits{S.}}:
Virtuous people and evil elites? the role of moralizing frames and normative distinctions in identifying populist discourse.
Political Research Exchange
(2024)
\end{botherref}
\endbibitem

\bibitem[\protect\citeauthoryear{Devlin et~al.}{2019}]{bert}
\begin{botherref}
\oauthor{\bsnm{Devlin}, \binits{J.}},
\oauthor{\bsnm{Chang}, \binits{M.-W.}},
\oauthor{\bsnm{Lee}, \binits{K.}},
\oauthor{\bsnm{Toutanova}, \binits{K.}}:
Bert: Pre-training of deep bidirectional transformers for language understanding.
Proceedings of NAACL-HLT,
4171--4186
(2019)
\end{botherref}
\endbibitem

\bibitem[\protect\citeauthoryear{Zhong et~al.}{2023}]{chatgpt_vs_bert}
\begin{botherref}
\oauthor{\bsnm{Zhong}, \binits{Q.}},
\oauthor{\bsnm{Ding}, \binits{L.}},
\oauthor{\bsnm{Liu}, \binits{J.}},
\oauthor{\bsnm{Du}, \binits{B.}},
\oauthor{\bsnm{Tao}, \binits{D.}}:
{Can ChatGPT Understand Too? A Comparative Study on ChatGPT and Fine-tuned BERT}.
arXiv:2302.10198
(2023)
\end{botherref}
\endbibitem

\bibitem[\protect\citeauthoryear{Wang et~al.}{2024}]{model_selection}
\begin{botherref}
\oauthor{\bsnm{Wang}, \binits{Y.}},
\oauthor{\bsnm{Qu}, \binits{W.}},
\oauthor{\bsnm{Ye}, \binits{X.}}:
{Selecting Between BERT and GPT for Text Classification in Political Science Research}.
arXiv:2411.05050
(2024)
\end{botherref}
\endbibitem

\bibitem[\protect\citeauthoryear{Wodak}{2015}]{Wodak2015}
\begin{bbook}
\bauthor{\bsnm{Wodak}, \binits{R.}}:
\bbtitle{The Politics of Fear: What Right-Wing Populist Discourses Mean}.
\bpublisher{SAGE Publications},
\blocation{London}
(\byear{2015}).
\burl{https://uk.sagepub.com/en-gb/eur/the-politics-of-fear/book246482}
\end{bbook}
\endbibitem

\bibitem[\protect\citeauthoryear{Moffitt}{2016}]{Moffitt2016}
\begin{bbook}
\bauthor{\bsnm{Moffitt}, \binits{B.}}:
\bbtitle{The Global Rise of Populism: Performance, Political Style, and Representation}.
\bpublisher{Stanford University Press},
\blocation{Stanford, CA}
(\byear{2016}).
\burl{https://www.sup.org/books/title/?id=24978}
\end{bbook}
\endbibitem

\bibitem[\protect\citeauthoryear{Judis}{2016}]{judis2016populist}
\begin{bbook}
\bauthor{\bsnm{Judis}, \binits{J.B.}}:
\bbtitle{The Populist Explosion: How the Great Recession Transformed American and European Politics}.
\bpublisher{Columbia Global Reports},
\blocation{New York, NY}
(\byear{2016})
\end{bbook}
\endbibitem

\bibitem[\protect\citeauthoryear{Le~Normand and Thai-Van}{2022}]{lenormandRoleFunctionWords2022}
\begin{botherref}
\oauthor{\bsnm{Le~Normand}, \binits{M.-T.}},
\oauthor{\bsnm{Thai-Van}, \binits{H.}}:
The role of {{Function Words}} to build syntactic knowledge in {{French-speaking}} children
\textbf{12}(1),
544
(2022)
\doiurl{10.1038/s41598-021-04536-6} .
Accessed 2025-05-02
\end{botherref}
\endbibitem

\bibitem[\protect\citeauthoryear{Lakoff}{1990}]{Lakoff1990}
\begin{bbook}
\bauthor{\bsnm{Lakoff}, \binits{R.T.}}:
\bbtitle{Talking Power: The Politics of Language in Our Lives}.
\bpublisher{Basic Books},
\blocation{New York}
(\byear{1990}).
\burl{https://archive.org/details/talkingpowerpoli00lako}
\end{bbook}
\endbibitem

\bibitem[\protect\citeauthoryear{Pennebaker et~al.}{2003}]{pennebaker2003psychological}
\begin{barticle}
\bauthor{\bsnm{Pennebaker}, \binits{J.W.}},
\bauthor{\bsnm{Mehl}, \binits{M.R.}},
\bauthor{\bsnm{Niederhoffer}, \binits{K.G.}}:
\batitle{Psychological aspects of natural language use: Our words, our selves}.
\bjtitle{Annual Review of Psychology}
\bvolume{54},
\bfpage{547}--\blpage{577}
(\byear{2003})
\doiurl{10.1146/annurev.psych.54.101601.145041}
\end{barticle}
\endbibitem

\bibitem[\protect\citeauthoryear{Lacatus and Meibauer}{2022}]{Corina}
\begin{barticle}
\bauthor{\bsnm{Lacatus}, \binits{C.}},
\bauthor{\bsnm{Meibauer}, \binits{G.}}:
\batitle{‘saying it like it is’: Right-wing populism, international politics, and the performance of authenticity}.
\bjtitle{The British Journal of Politics and International Relations}
\bvolume{24}(\bissue{3}),
\bfpage{437}--\blpage{457}
(\byear{2022})
\doiurl{10.1177/13691481221089137}
\end{barticle}
\endbibitem

\bibitem[\protect\citeauthoryear{Pappas and Kriesi}{2015}]{EuropeanPopulism2015}
\begin{bbook}
\bauthor{\bsnm{Pappas}, \binits{T.}},
\bauthor{\bsnm{Kriesi}, \binits{H.}}:
\bbtitle{European Populism in the Shadow of the Great Recession}.
\bpublisher{ECPR Press},
\blocation{Colchester}
(\byear{2015})
\end{bbook}
\endbibitem

\bibitem[\protect\citeauthoryear{Hawkins}{2009}]{hawkins2009ischavez}
\begin{barticle}
\bauthor{\bsnm{Hawkins}, \binits{K.A.}}:
\batitle{Is chávez populist?}
\bjtitle{Comparative Political Studies}
\bvolume{42}(\bissue{8}),
\bfpage{1040}--\blpage{1067}
(\byear{2009})
\doiurl{10.1177/0010414009331721}
\end{barticle}
\endbibitem

\bibitem[\protect\citeauthoryear{Bonikowski and Gidron}{2015}]{10.1093/sf/sov120}
\begin{barticle}
\bauthor{\bsnm{Bonikowski}, \binits{B.}},
\bauthor{\bsnm{Gidron}, \binits{N.}}:
\batitle{The populist style in american politics: Presidential campaign discourse, 1952–1996}.
\bjtitle{Social Forces}
\bvolume{94}(\bissue{4}),
\bfpage{1593}--\blpage{1621}
(\byear{2015})
\doiurl{10.1093/sf/sov120}
{\href{https://arxiv.org/abs/https://academic.oup.com/sf/article-pdf/94/4/1593/7572409/sov120.pdf}{{https://academic.oup.com/sf/article-pdf/94/4/1593/7572409/sov120.pdf}}}
\end{barticle}
\endbibitem

\bibitem[\protect\citeauthoryear{Norris and Inglehart}{2019}]{norris2019cultural}
\begin{bbook}
\bauthor{\bsnm{Norris}, \binits{P.}},
\bauthor{\bsnm{Inglehart}, \binits{R.}}:
\bbtitle{Cultural Backlash: Trump, Brexit, and Authoritarian Populism}.
\bpublisher{Cambridge University Press},
\blocation{Cambridge}
(\byear{2019})
\end{bbook}
\endbibitem

\bibitem[\protect\citeauthoryear{Hawkins et~al.}{2019}]{gpd}
\begin{botherref}
\oauthor{\bsnm{Hawkins}, \binits{K.A.}},
\oauthor{\bsnm{Aguilar}, \binits{R.}},
\oauthor{\bsnm{Castanho~Silva}, \binits{B.}},
\oauthor{\bsnm{Jenne}, \binits{E.K.}},
\oauthor{\bsnm{Kocijan}, \binits{B.}},
\oauthor{\bsnm{Rovira~Kaltwasser}, \binits{C.}}:
Measuring populist discourse: The global populism database.
Harvard Dataverse
(2019)
\end{botherref}
\endbibitem

\bibitem[\protect\citeauthoryear{Bonikowski et~al.}{2022}]{politics_as_usual}
\begin{botherref}
\oauthor{\bsnm{Bonikowski}, \binits{B.}},
\oauthor{\bsnm{Luo}, \binits{Y.}},
\oauthor{\bsnm{Stuhler}, \binits{O.}}:
{Politics as Usual? Measuring Populism, Nationalism, and Authoritarianism in U.S. Presidential Campaigns (1952–2020) with Neural Language Models}.
Sociological Methods \& Research
(2022)
\end{botherref}
\endbibitem

\end{thebibliography}

\newpage
\section*{Appendix}

\section*{Model Comparison between Google Translate and GPT-4o} 

\label{google}

Table~\ref{google} presents a comparison between models fine-tuned on English translations of German texts from Google Translate and those using translations from GPT-4o. The results indicate that models trained on GPT-4o translations perform slightly better, with an improvement of up to one percentage point in both micro and macro averages.

\begin{table}[h!]
\renewcommand{\arraystretch}{1.1}
\centering
\setlength{\tabcolsep}{15pt}
\begin{tabular}{|l|c|c|}
\hline
  Dimension & \multicolumn{1}{c|}{Google Translate}  & \multicolumn{1}{c|}{GPT-4o} \\ \hline
Anti-Elitism                        & 0.84    & \textbf{0.84} \\ \hline
People-Centrism                     & 0.70    & \textbf{0.72} \\ \hline
Left-Wing                  & 0.71    & \textbf{0.71} \\ \hline
Right-Ring                 & 0.72    & \textbf{0.73} \\ \hline
Micro Average              & 0.77    & \textbf{0.78} \\ \hline
Macro Average              & 0.74    & \textbf{0.75} \\ \hline

\end{tabular}
\caption{F1 scores across different dimensions for UniPop using translations from Google Translate and from GPT-4o. All results are averaged over three random runs. Better or equal results (higher F1 score) in bold.}
\label{google}
\end{table}

\section*{Descriptive Statistics of the LIWC Features (94 in total)} 

\begin{longtable}{p{1.4cm}lrrrrrrrr}
\hline\hline
Feature & Count & Mean & STD & Min & 25\% & 50\% & 75\% & Max \\
\hline
\endfirsthead

\hline
Feature & Count & Mean & STD & Min & 25\% & 50\% & 75\% & Max \\
\hline
\endhead

\hline
\endfoot

\endlastfoot

Segment & 308 & 1.00 & 0.00 & 1.00 & 1.00 & 1.00 & 1.00 & 1.00 \\
WC & 308 & 6966.12 & 5817.26 & 135.00 & 2875.00 & 5204.00 & 9036.00 & 34035.00 \\
Analytic & 308 & 90.18 & 9.04 & 57.85 & 86.80 & 93.72 & 96.50 & 99.00 \\
Clout & 308 & 70.91 & 12.84 & 38.33 & 59.89 & 69.37 & 82.11 & 96.72 \\
Authentic & 308 & 21.69 & 9.97 & 2.49 & 14.54 & 18.46 & 27.72 & 50.14 \\
Tone & 308 & 68.17 & 16.06 & 9.46 & 59.46 & 67.75 & 78.68 & 99.00 \\
WPS & 308 & 29.28 & 9.39 & 11.84 & 20.92 & 29.05 & 36.16 & 62.73 \\
Sixltr & 308 & 25.53 & 3.20 & 15.59 & 23.49 & 25.89 & 27.62 & 33.80 \\
Dic & 308 & 84.01 & 2.95 & 76.54 & 81.94 & 83.99 & 86.10 & 90.55 \\
function & 308 & 52.48 & 3.23 & 41.22 & 50.39 & 53.16 & 54.95 & 59.45 \\
pronoun & 308 & 10.23 & 2.41 & 4.98 & 8.27 & 10.00 & 11.91 & 17.35 \\
ppron & 308 & 5.15 & 2.15 & 1.83 & 3.31 & 4.86 & 6.82 & 11.79 \\
i & 308 & 0.97 & 0.77 & 0.07 & 0.51 & 0.79 & 1.16 & 6.67 \\
we & 308 & 2.56 & 1.76 & 0.00 & 1.06 & 2.01 & 3.88 & 7.95 \\
you & 308 & 0.30 & 0.35 & 0.00 & 0.06 & 0.17 & 0.42 & 1.84 \\
shehe & 308 & 0.31 & 0.27 & 0.00 & 0.13 & 0.26 & 0.40 & 2.08 \\
they & 308 & 1.01 & 0.43 & 0.00 & 0.70 & 0.97 & 1.26 & 3.55 \\
ipron & 308 & 5.08 & 0.90 & 2.42 & 4.45 & 5.04 & 5.68 & 8.11 \\
article & 308 & 9.67 & 1.88 & 5.98 & 8.17 & 9.66 & 11.35 & 13.98 \\
prep & 308 & 17.16 & 2.06 & 12.28 & 15.63 & 17.44 & 18.91 & 20.90 \\
auxverb & 308 & 7.43 & 0.99 & 4.75 & 6.73 & 7.34 & 8.02 & 11.94 \\
adverb & 308 & 2.71 & 0.60 & 1.41 & 2.31 & 2.58 & 3.06 & 5.08 \\
conj & 308 & 5.96 & 0.83 & 3.70 & 5.37 & 5.83 & 6.45 & 9.56 \\
negate & 308 & 1.08 & 0.41 & 0.00 & 0.81 & 1.02 & 1.25 & 2.58 \\
verb & 308 & 10.80 & 1.84 & 6.91 & 9.46 & 10.49 & 11.89 & 17.50 \\
adj & 308 & 4.65 & 0.77 & 0.74 & 4.12 & 4.58 & 5.18 & 6.68 \\
compare & 308 & 2.32 & 0.48 & 0.00 & 2.00 & 2.25 & 2.57 & 4.16 \\
interrog & 308 & 1.36 & 0.43 & 0.47 & 1.05 & 1.32 & 1.67 & 3.07 \\
number & 308 & 1.68 & 1.12 & 0.00 & 0.81 & 1.54 & 2.24 & 6.11 \\
quant & 308 & 2.28 & 0.46 & 1.15 & 1.96 & 2.24 & 2.57 & 4.21 \\
affect & 308 & 5.90 & 1.37 & 3.71 & 4.81 & 5.60 & 6.82 & 10.15 \\
posemo & 308 & 4.11 & 1.05 & 2.05 & 3.32 & 3.87 & 4.75 & 7.83 \\
negemo & 308 & 1.73 & 0.66 & 0.46 & 1.29 & 1.58 & 2.00 & 4.10 \\
anx & 308 & 0.30 & 0.16 & 0.00 & 0.20 & 0.27 & 0.36 & 1.25 \\
anger & 308 & 0.61 & 0.44 & 0.07 & 0.35 & 0.49 & 0.70 & 2.91 \\
sad & 308 & 0.31 & 0.15 & 0.00 & 0.22 & 0.27 & 0.36 & 1.15 \\
social & 308 & 7.74 & 2.66 & 3.79 & 5.58 & 7.05 & 9.62 & 14.47 \\
family & 308 & 0.11 & 0.16 & 0.00 & 0.02 & 0.05 & 0.12 & 0.83 \\
friend & 308 & 0.17 & 0.12 & 0.00 & 0.09 & 0.15 & 0.22 & 0.74 \\
female & 308 & 0.11 & 0.14 & 0.00 & 0.02 & 0.06 & 0.14 & 0.81 \\
male & 308 & 0.46 & 0.33 & 0.02 & 0.24 & 0.39 & 0.56 & 2.54 \\
cogproc & 308 & 9.88 & 1.36 & 6.83 & 9.02 & 9.71 & 10.81 & 14.44 \\
insight & 308 & 1.81 & 0.42 & 0.74 & 1.51 & 1.79 & 2.06 & 3.21 \\
cause & 308 & 1.85 & 0.39 & 0.00 & 1.64 & 1.85 & 2.10 & 3.39 \\
discrep & 308 & 1.43 & 0.47 & 0.45 & 1.09 & 1.35 & 1.74 & 3.03 \\
tentat & 308 & 1.81 & 0.51 & 0.56 & 1.46 & 1.80 & 2.09 & 4.22 \\
certain & 308 & 1.84 & 0.54 & 0.98 & 1.40 & 1.72 & 2.11 & 3.92 \\
differ & 308 & 2.51 & 0.63 & 1.04 & 2.08 & 2.48 & 2.84 & 5.11 \\
percept & 308 & 0.85 & 0.40 & 0.10 & 0.60 & 0.74 & 1.02 & 2.82 \\
see & 308 & 0.35 & 0.18 & 0.00 & 0.24 & 0.31 & 0.40 & 1.27 \\
hear & 308 & 0.21 & 0.19 & 0.00 & 0.09 & 0.15 & 0.27 & 1.22 \\
feel & 308 & 0.19 & 0.12 & 0.00 & 0.11 & 0.16 & 0.24 & 1.11 \\
bio & 308 & 0.83 & 0.37 & 0.22 & 0.55 & 0.74 & 1.06 & 2.34 \\
body & 308 & 0.21 & 0.14 & 0.00 & 0.13 & 0.17 & 0.25 & 0.91 \\
health & 308 & 0.49 & 0.30 & 0.00 & 0.29 & 0.41 & 0.65 & 1.88 \\
sexual & 308 & 0.03 & 0.04 & 0.00 & 0.00 & 0.02 & 0.04 & 0.28 \\
ingest & 308 & 0.12 & 0.10 & 0.00 & 0.07 & 0.10 & 0.15 & 0.74 \\
drives & 308 & 11.42 & 2.82 & 7.08 & 8.92 & 10.90 & 13.87 & 17.84 \\
affiliation & 308 & 3.63 & 2.08 & 0.00 & 1.78 & 3.06 & 5.39 & 9.19 \\
achieve & 308 & 2.20 & 0.68 & 1.10 & 1.64 & 2.06 & 2.73 & 3.99 \\
power & 308 & 4.61 & 0.73 & 1.77 & 4.10 & 4.62 & 5.07 & 6.67 \\
reward & 308 & 1.26 & 0.37 & 0.43 & 1.00 & 1.21 & 1.46 & 2.85 \\
risk & 308 & 0.96 & 0.28 & 0.00 & 0.77 & 0.94 & 1.13 & 1.85 \\
focuspast & 308 & 2.44 & 0.74 & 0.59 & 1.91 & 2.47 & 2.90 & 4.72 \\
focuspresent & 308 & 7.05 & 1.74 & 3.59 & 5.75 & 6.77 & 8.22 & 12.74 \\
focusfuture & 308 & 1.56 & 0.52 & 0.58 & 1.17 & 1.50 & 1.88 & 4.00 \\
relativ & 308 & 12.49 & 1.97 & 7.36 & 11.17 & 12.16 & 14.12 & 17.17 \\
motion & 308 & 1.36 & 0.46 & 0.34 & 1.04 & 1.29 & 1.70 & 3.34 \\
space & 308 & 7.01 & 0.91 & 3.93 & 6.46 & 7.02 & 7.63 & 9.50 \\
time & 308 & 4.17 & 1.13 & 1.53 & 3.37 & 4.07 & 4.85 & 7.79 \\
work & 308 & 4.91 & 1.49 & 1.33 & 3.92 & 4.76 & 5.78 & 10.22 \\
leisure & 308 & 0.29 & 0.17 & 0.00 & 0.19 & 0.26 & 0.35 & 1.48 \\
home & 308 & 0.26 & 0.18 & 0.00 & 0.14 & 0.20 & 0.34 & 1.06 \\
money & 308 & 1.61 & 0.84 & 0.00 & 1.02 & 1.48 & 2.09 & 4.89 \\
relig & 308 & 0.34 & 0.28 & 0.00 & 0.16 & 0.26 & 0.40 & 1.95 \\
death & 308 & 0.25 & 0.23 & 0.00 & 0.13 & 0.20 & 0.31 & 1.71 \\
informal & 308 & 0.15 & 0.10 & 0.00 & 0.10 & 0.13 & 0.19 & 0.73 \\
swear & 308 & 0.01 & 0.01 & 0.00 & 0.00 & 0.00 & 0.01 & 0.08 \\
netspeak & 308 & 0.01 & 0.03 & 0.00 & 0.00 & 0.00 & 0.01 & 0.28 \\
assent & 308 & 0.04 & 0.05 & 0.00 & 0.00 & 0.02 & 0.05 & 0.62 \\
nonflu & 308 & 0.10 & 0.07 & 0.00 & 0.06 & 0.09 & 0.13 & 0.42 \\
filler & 308 & 0.00 & 0.00 & 0.00 & 0.00 & 0.00 & 0.00 & 0.01 \\
AllPunc & 308 & 10.89 & 2.68 & 6.16 & 8.96 & 10.25 & 12.18 & 22.40 \\
Period & 308 & 3.84 & 1.46 & 1.42 & 2.73 & 3.42 & 4.85 & 15.35 \\
Comma & 308 & 5.11 & 1.00 & 2.43 & 4.44 & 5.12 & 5.72 & 9.09 \\
Colon & 308 & 0.12 & 0.15 & 0.00 & 0.03 & 0.06 & 0.17 & 1.48 \\
SemiC & 308 & 0.28 & 0.21 & 0.00 & 0.13 & 0.22 & 0.38 & 1.36 \\
QMark & 308 & 0.04 & 0.08 & 0.00 & 0.00 & 0.00 & 0.06 & 0.60 \\
Exclam & 308 & 0.01 & 0.04 & 0.00 & 0.00 & 0.00 & 0.00 & 0.46 \\
Dash & 308 & 0.71 & 0.74 & 0.00 & 0.31 & 0.46 & 0.76 & 5.13 \\
Quote & 308 & 0.14 & 0.17 & 0.00 & 0.02 & 0.09 & 0.20 & 0.82 \\
Apostro & 308 & 0.37 & 0.64 & 0.00 & 0.04 & 0.11 & 0.36 & 3.55 \\
Parenth & 308 & 0.05 & 0.24 & 0.00 & 0.00 & 0.00 & 0.04 & 3.83 \\
OtherP & 308 & 0.22 & 0.23 & 0.00 & 0.02 & 0.17 & 0.34 & 1.87 \\\hline\hline
% ...
\caption{Descriptive statistics of LIWC features extracted from U.S. presidential election speeches.} \\
\label{liwc}
\end{longtable}

\end{document}